\documentclass[conference]{IEEEtran}

\usepackage{geometry}
\usepackage{amsmath,amssymb,amsfonts}
\usepackage{algorithmic}
\usepackage{graphicx, subfigure}
\usepackage{multirow, booktabs, makecell, threeparttable, arydshln}
\usepackage{textcomp}
\usepackage{xcolor}
\usepackage[hyphens]{url}
\usepackage{hyperref}
\usepackage[capitalise]{cleveref}
\usepackage[
  style=ieee,
  backend=biber,
  maxnames=6,
  doi=false,
  url=false,
  isbn=false
]{biblatex}
\begin{document}

\newgeometry{top=1in,left=0.75in,right=0.75in,bottom=0.75in}

\title{Trainable Smooth-Rotation Transforms with Learned Channel Scales for LLM Quantization}

\author{%
  \begin{minipage}[t]{0.32\linewidth}
    \centering
    Patrik Czakó\\[0.5ex]
    \textit{Doctoral School of Applied Informatics and Applied Mathematics, Obuda University}\\
    Budapest, Hungary\\
    czako.patrik@stud.uni-obuda.hu
  \end{minipage}\hfill
  \begin{minipage}[t]{0.32\linewidth}
    \centering
    Gábor Kertész\\[0.5ex]
    \textit{John von Neumann Faculty of Informatics, Obuda University}\\
    Budapest, Hungary\\
    kertesz.gabor@nik.uni-obuda.hu
  \end{minipage}\hfill
  \begin{minipage}[t]{0.32\linewidth}
    \centering
    Sándor Szénási\\[0.5ex]
    \textit{John von Neumann Faculty of Informatics, Obuda University}\\
    Budapest, Hungary\\
    szenasi.sandor@nik.uni-obuda.hu
  \end{minipage}%
}

\maketitle

\begin{abstract}
Post-training quantization (PTQ) is one of the most practical ways to reduce the serving cost of Large Language Models (LLMs), but activation quantization remains difficult because outlier-dominated channels lead to large quantization errors.
This paper investigates whether part of this degradation is caused by over-migration in scaling-based equivalent transformations.
We introduce a quantile-robust scaling policy for SmoothRot-style transforms by replacing max-based activation statistics with high quantiles, and we complement it with constrained gradient-based optimization of channel scales.
On LLaMA-3.2-1B under W4A4 quantization, quantile-only policy search improves selected-layer error by 11.1\% over the SmoothRot baseline, joint \((\alpha,q)\) search improves it by 12\%, and training reaches 18.5\%.
Replaying the best selected-layer policy on all decoder-block down-projection layers reduces the corresponding full-layer mean error from 97.51 to 78.08 (19.9\%). The results show that robust migration control and lightweight scale learning provide consistent gains over max-based fixed policies while preserving the equivalent-transform framework.

\end{abstract}


\section{Introduction}

Large Language Models (LLMs) have made substantial progress in generative and reasoning tasks, but their deployment still faces strong memory and latency constraints \cite{onozo2025comparative}.
Post-training quantization (PTQ) is therefore a central optimization path because it does not require costly full-model retraining \cite{dettmers2022llm,xiao2023smoothquant}.
In low-bit settings, however, activation quantization remains significantly harder than weight quantization due to emergent outliers \cite{dettmers2022llm,ashkboos2024quarot,yang2024mitigating}.

Equivalent transformations are a widely used way to mitigate this problem.
Channel-wise scaling can reduce activation range but may transfer quantization difficulty to weights \cite{xiao2023smoothquant}.
Rotation redistributes large values across channels and is effective in many cases \cite{ashkboos2024quarot,liu2024spinquant}.
Our previous studies examined these mechanisms in detail and then combined them in the SmoothRot architecture \cite{czako2025turning,czako2025smoothrot}.
In particular, we observed that the scaling component can be beneficial, but can migrate quantization difficulty to weights too aggressively leading to suboptimal performance.

Building on this observation, the present work focuses on controlling migration strength more robustly and testing whether this resolves a key failure mode of max-based scaling.
Our contributions are threefold:
(i) we generalize the SmoothQuant-style scale formula \cite{xiao2023smoothquant}  with a quantile parameter that reduces sensitivity to rare activation spikes,
(ii) we introduce a policy-structured optimization protocol that separates no-training and training regimes,
and (iii) we provide controlled evidence on LLaMA-3.2-1B \cite{grattafiori2024llama3} (W4A4) that this strategy yields consistent gains from selected layers to full-network replay.
The central finding is that robust scale estimation contributes more than alpha-only retuning in the no-training regime, while constrained scale learning provides the strongest final improvements.

\section{Related Work}

\label{sec:related_work}

Low-bit post-training quantization has progressed rapidly, but a persistent gap remains between weight-only and weight-activation quantization quality in LLMs \cite{kim2024squeezellm,tseng2024quip,sun2025flat,czako2025activation}. The main reason is that activation distributions are highly non-uniform and contain outliers that increase quantization errors significantly \cite{czako2025activation,xiao2023smoothquant,yang2024mitigating,sun2024massive}.

Equivalent transformations are a dominant mitigation family. Channel-wise scaling (smoothing) reduces activation range by rebalancing activation and weight magnitudes \cite{xiao2023smoothquant,shao2024omniquant}. Rotation-based methods use orthogonal transforms (often Hadamard-structured) to spread large values across channels \cite{ashkboos2024quarot,liu2024spinquant,lin2024duquant}. While both approaches are effective in many regimes, their behavior differs for systematic versus massive outliers \cite{czako2025activation}, and the best strategy depends on model/layer characteristics and quantization scheme.

Previous work \cite{czako2025turning} analyzed layer-wise quantization error formation and proposed combining channel-wise scaling with rotation to balance their complementary strengths. The paper provides mathematical intuition showing why smoothing can help rotation in extreme outlier regimes by partially redistributing difficulty before orthogonal mixing. SmoothRot \cite{czako2025smoothrot} integrates this idea into a full LLM architecture and demonstrates end-to-end gains in W4A4KV4 settings on multiple models, while preserving the low-latency profile of rotation-only techniques.

The present paper addresses a narrower but critical unresolved question from this line of research: \emph{is degradation sometimes caused by over-migration from scaling rather than by transformation choice alone?} We therefore focus on policy-structured optimization of the scaling strength and introduce a quantile-based refinement of previously used scaling factor formula. Also, we propose gradient-guided optimization of these scaling parameters in a controlled training framework.

\section{Method}

\label{sec:method}

\subsection{Problem formulation}

For a linear projection layer with activation tensor \(\mathbf{X}\in\mathbb{R}^{n\times c_{\mathrm{in}}}\) and weight matrix
\(\mathbf{W}\in\mathbb{R}^{c_{\mathrm{in}}\times c_{\mathrm{out}}}\), the layer output is $\mathbf{Y} = \mathbf{X}\mathbf{W}$.
Here, \(n\) is the sequence length, while \(c_{\mathrm{in}}\) and \(c_{\mathrm{out}}\) denote input and output channel counts, respectively.
In this paper, following SmoothRot, we apply the equivalent transform only to the \emph{down-projection} linear layers in each decoder block, where extreme activation outliers predominantly occur. Other projections are left unchanged and are out of scope.
Let \(Q_b(\cdot)\) denote \(b\)-bit quantization. The quantized output is \(\hat{\mathbf{Y}} = Q_b(\mathbf{X})Q_b(\mathbf{W})\).
We use the linear output error
\begin{equation}
    \label{eq:linear_output_error}
    E\!\left(\mathbf{Y},\hat{\mathbf{Y}}\right)=\left\|\mathbf{Y}-\hat{\mathbf{Y}}\right\|_{F},
\end{equation}
and its normalized variant
\begin{equation}
    \tilde{E}\!\left(\mathbf{Y},\hat{\mathbf{Y}}\right)=\frac{E\!\left(\mathbf{Y},\hat{\mathbf{Y}}\right)}{\left\|\mathbf{Y}\right\|_{F}+\varepsilon},
\end{equation}
where \(\varepsilon\) is a small constant for numerical stability. Equivalent-transform methods aim to find an invertible linear transformation that reduces these errors under low-bit quantization by mitigating activation outliers.
The matrix of this transformation is denoted by \(\mathbf{A}\in\mathbb{R}^{c_{\mathrm{in}}\times c_{\mathrm{in}}}\). Applying \(\mathbf{A}\) to the input and its inverse to the weights yields an equivalent representation:
\begin{equation} \label{eq:eq_transformation}
\mathbf{Y} = \mathbf{X} \mathbf{W} = \mathbf{X}\underbrace{\left( \mathbf{A}\mathbf{A}^{-1}\right)}_{\mathbb{I}} \mathbf{W} = \underbrace{\left(\mathbf{X} \mathbf{A}\right)}_{\hat{\mathbf{X}}} \cdot \underbrace{\left(\mathbf{A}^{-1} \mathbf{W}\right)}_{\hat{\mathbf{W}}}.
\end{equation}
The objective is to choose \(\mathbf{A}\) such that quantization error of
\((\hat{\mathbf{X}},\hat{\mathbf{W}})\) is reduced under low-bit symmetric quantization.

\subsection{Parameterized transformation}

We use a SmoothRot-style parameterization with channel-wise smoothing and orthogonal rotation. The transformation matrix is constructed as
\begin{equation}
\mathbf{A}=\left(\mathbf{H}\mathbf{\Lambda}\right)^{-1},
\end{equation}
\label{eq:a_h_lambda}
where \(\mathbf{H}\) is an orthogonal Hadamard matrix, and \(\mathbf{\Lambda}=\mathrm{diag}(\mathbf{s})\) is a diagonal matrix with positive entries, constructed from the so-called scaling factor $\mathbf{s} \in \mathbb{R}^{c_{in}}$.
This ordering applies smoothing first and then rotates activations through the equivalent transform in~\eqref{eq:eq_transformation}.

Previous studies \cite{xiao2023smoothquant,czako2025smoothrot} obtained \(\mathbf{s}\) using the channel-wise max of both activations and weights, with a fixed migration strength \(\alpha\in[0,1]\) that controls the balance between them.
However, this heuristic is sensitive to activation outliers, leading to suboptimal scaling. We therefore introduce a more robust formula that generalizes SmoothQuant \cite{xiao2023smoothquant} by replacing max-based statistics with quantiles:
\begin{equation}
s_j(\alpha,q)=\frac{\mathcal{Q}_{q}\!\left(|\mathbf{X}_{:,j}|\right)^{\alpha}}{\left(\max_{k}|\mathbf{W}_{j,k}|\right)^{1-\alpha}},
\label{eq:warmstart_scale}
\end{equation}
where \(\mathcal{Q}_{q}(\cdot)\) is the empirical quantile at level \(q\). For \(q\to 1\), this approaches max-based calibration; keeping \(q<1\) makes the estimate less dominated by rare outliers.

\subsection{Gradient-based optimization}

Prior work \cite{shao2024omniquant} has shown that learnable smoothing can further reduce quantization error. Following this line, we also include experiment families that optimize \(\mathbf{s}\) by gradient descent.
This optimization, however, is challenging due to the non-differentiability of quantization, thus we use a straight-through estimator (STE) \cite{bengio2013estimating} to optimize \(\mathbf{s}\) by backpropagating through the quantization operation.
Specifically, we optimize a straight-through surrogate objective using the normalized error for scale invariance:
\begin{equation}
\mathcal{L}_{\mathrm{STE}} = \frac{1}{|\mathcal{S}|}\sum_{\ell\in\mathcal{S}}
\tilde{E}\!\left(\mathbf{X}_{\ell}\mathbf{W}_{\ell},\;Q_b(\hat{\mathbf{X}}_{\ell})Q_b(\hat{\mathbf{W}}_{\ell})\right).
\end{equation}
where \(\ell\) indexes decoder blocks and \(\mathcal{S}\) is either the selected-layer set or the full set of decoder-block down-projection layers, depending on family.

\section{Experimental Setup}

\label{sec:experimental_setup}

This study focuses on reaching the limits of SmoothRot-based quantization in terms of achievable accuracy preservation. To this end, we present a detailed experimental protocol designed to systematically evaluate and optimize our proposed technique.

\subsection{Baseline and constraints}

We evaluate the proposed methods against SmoothRot in our main experiments, also including QuaRot as a rotation-only baseline to see the extra improvement from optimized scaling.
All reported values are reproduced from the same set of runs, ensuring a fair comparison. QuaRot is implemented by setting \(\mathbf{s}=\mathbf{1}\), whereas SmoothRot uses the original heuristic max-based scaling (\(q=1.0\)).
Although our protocol can be applied to various models and architectures, we focus on a single model and quantization setting to maintain a controlled environment for analysis.
We choose the 1B-parameter variant of LLaMA-3.2 \cite{grattafiori2024llama3} as it contains the same architectural building blocks as larger LLMs and also suffers from the activation outlier problem, while enabling faster experimentation. The quantization setting is fixed to 4-bit weights and activations (W4A4), which is a common setting for efficient inference.
For training and evaluation, we use the Wikitext-2 dataset \cite{merity2017pointer} as it is a standard benchmark for language modeling tasks and provides a large and diverse set of text data. In total, we take 256 samples of 128 tokens, and split them into separate train and test sets with an 80/20 ratio.

\subsection{Deterministic layer selection}

Following \cite{czako2025turning}, all baselines and policies in this paper are evaluated on the down-projection linear layers in each decoder block.
Extreme activation outliers predominantly appear in these down-projection layers, and we therefore restrict ranking, selection, and optimization to them.

Our protocol first runs a full-layer baseline using the original SmoothRot transformation (\(\alpha=0.5, q=1.0\)) on the test set and ranks decoder blocks by the down-projection linear output error defined in~\eqref{eq:linear_output_error}.
A fixed selected-layer set \(\mathcal{S}_{\mathrm{sel}}\) is then formed as top-3 critical layers with highest error plus 2 normal layers.

\subsection{Experiment families}

The orchestration flow consists of baseline capture/ranking, selected-layer policy derivation, no-training sweeps, training sweeps, and full-layer replay.
The protocol is organized into no-training families (N1--N3) and training families (T1--T3), where the exact family configurations are:

\paragraph{N1: max-based scaling sensitivity in \(\alpha\)}
N1 tests the SmoothQuant heuristic scaling under max-based statistics (\(q=1\)) while varying migration strength \(\alpha\):
\begin{equation}
    \begin{split}
        \min_{\alpha\in\mathcal{A}_{\mathrm{N1}}}\;\frac{1}{|\mathcal{S}_{\mathrm{sel}}|}\sum_{\ell\in\mathcal{S}_{\mathrm{sel}}}
        E_{\ell}(\alpha, q=1),\\
        \alpha_i = 0.25 + 0.025i,\quad i=0,\dots,20,
    \end{split}
\end{equation}
where \(E_{\ell}(\alpha,q)\) denotes the linear output error after applying the corresponding equivalent transform and quantization.

\paragraph{N2: quantile-robust scaling in \(q\)}
N2 replaces max-based activation statistics with quantiles to reduce sensitivity to rare spikes:
\begin{equation}
\min_{q\in\mathcal{Q}_{\mathrm{N2}}}\;\frac{1}{|\mathcal{S}_{\mathrm{sel}}|}\sum_{\ell\in\mathcal{S}_{\mathrm{sel}}}
E_{\ell}(\alpha=0.5, q),
\end{equation}
with scaling given by~\eqref{eq:warmstart_scale}. As \(q\) increases toward one, behavior approaches max-based calibration.
We sweep over 16 quantiles from the range \([0.95,1.0]\) with log spacing to capture the steep error increase near \(q=1\).

\paragraph{N3: joint \((\alpha,q)\) policy search}
N3 performs a compact grid search over candidates transferred from N1 and N2:
\begin{equation}
\min_{(\alpha,q)\in\mathcal{G}_{\mathrm{N3}}}\;\frac{1}{|\mathcal{S}_{\mathrm{sel}}|}\sum_{\ell\in\mathcal{S}_{\mathrm{sel}}}
E_{\ell}(\alpha,q), \quad
\mathcal{G}_{\mathrm{N3}}=\mathcal{A}^{\star}_{5}\times\mathcal{Q}^{\star}_{5},
\end{equation}
where \(\mathcal{A}^{\star}_{5}\) is the top-5 set from N1 and \(\mathcal{Q}^{\star}_{5}\) is the top-5 set from N2 with forced inclusion of \(q=1.0\).
Its intuition is that \(\alpha\) controls activation-weight migration strength, while \(q\) controls robustness of activation statistics, so jointly tuning both might yield better results.

\paragraph{T1: training from rotation-only initialization}
T1 starts from \(\mathbf{s}^{(0)}=\mathbf{1}\), i.e., rotation-only initialization, and the goal is to evaluate how learning-rate choice affects optimization stability and final error.
For the optimization, we use AdamW \cite{loshchilov2019decoupled} with a cosine decay schedule and no weight decay. The total number of steps is \(T=800\) with a warmup of \(T_w=5\) steps and early stopping if no improvement is seen in 80 steps. We sweep over 5 learning rates from logspace in the range \([5\times10^{-3}, 2\times10^{-2}]\).

\paragraph{T2: training from N3 warmstart policies}
T2 reuses the best N3 warmstart policies and trains with two learning rates: the best rate found by T1 and its immediate smaller neighbor.
Formally, initialization and optimization spaces are defined as
\begin{equation}
    \begin{split}
        \mathbf{s}^{(0)} &= \mathbf{s}(\alpha^{\star}, q^{\star}),\\
        (\alpha^{\star},q^{\star})&\in\mathcal{G}_{\mathrm{N3}}^{\star},\\
        \mathcal{G}_{\mathrm{T2}}&=\{(q_r,\alpha_r)\}_{r=1}^{5}\times\{\eta_{\mathrm{best}},\eta_{\mathrm{best-1}}\},
    \end{split}
\end{equation}
followed by the same optimization procedure as T1. This family tests whether the best no-training policies can further improve with training, and whether the best T1 learning rate remains optimal when combined with N3-informed warmstarts.

\paragraph{T3: full-network replay}
T3 replays the best T2 configuration on all decoder-block down-projection layers:
\begin{equation}
\mathcal{S}_{\mathrm{T3}} = \mathcal{S}_{\mathrm{all}},
\end{equation}
to verify that selected-layer conclusions transfer to full-network behavior, and to see the final end-to-end improvement over the SmoothRot and QuaRot baselines.

\section{Results}

\label{sec:results}

\begin{table}[t]
\centering
\caption{Selected critical and normal down-projection layers from baseline ranking.}
\label{tab:selected_layers}
\begin{tabular}{lcc}
\toprule
Layer type & Layer id & Baseline error \\
\midrule
Critical & 15 & 306.03 \\
Critical & 14 & 146.35 \\
Critical & 13 & 122.81 \\
Normal & 7 & 71.97 \\
Normal & 4 & 64.96 \\
\bottomrule
\end{tabular}
\end{table}

\subsection{Layer selection summary}

The baseline ranking selects three critical and two normal down-projection layers. The selected-layer baseline mean linear output error on the test set is 142.42, while the baseline mean over all down-projection layers is 97.51. The selected layers and baseline errors are shown in Table~\ref{tab:selected_layers}.

\begin{figure}[t]
    \centering
    \includegraphics[width=\columnwidth]{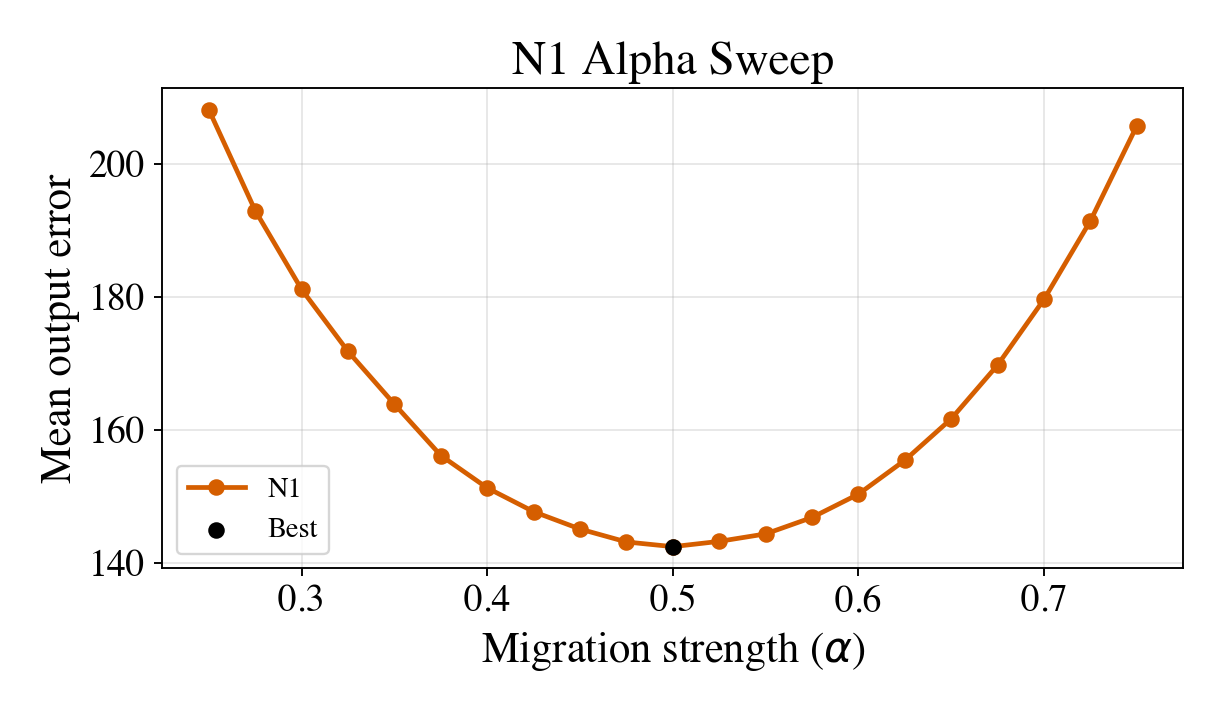}
    \caption{N1 alpha sweep over selected-layer mean linear output error.}
    \label{fig:n1_alpha_sweep}
\end{figure}

\subsection{Alpha Sweep (N1)}

Figure~\ref{fig:n1_alpha_sweep} shows the N1 alpha sweep results.
The best alpha is 0.5, which corresponds to the original SmoothQuant heuristic.
The results show a pronounced U-curve with degradation toward both ends.
This finding is consistent with the results of SmoothRot, i.e., both under-migration and over-migration can lead to suboptimal quantization error.
This experiment indicates that alpha-only optimization does not improve over the original SmoothRot heuristic, and motivates exploring more robust scaling methods that are less sensitive to outliers, such as quantile-based scaling in N2.

\begin{figure}[t]
    \centering
    \includegraphics[width=\columnwidth]{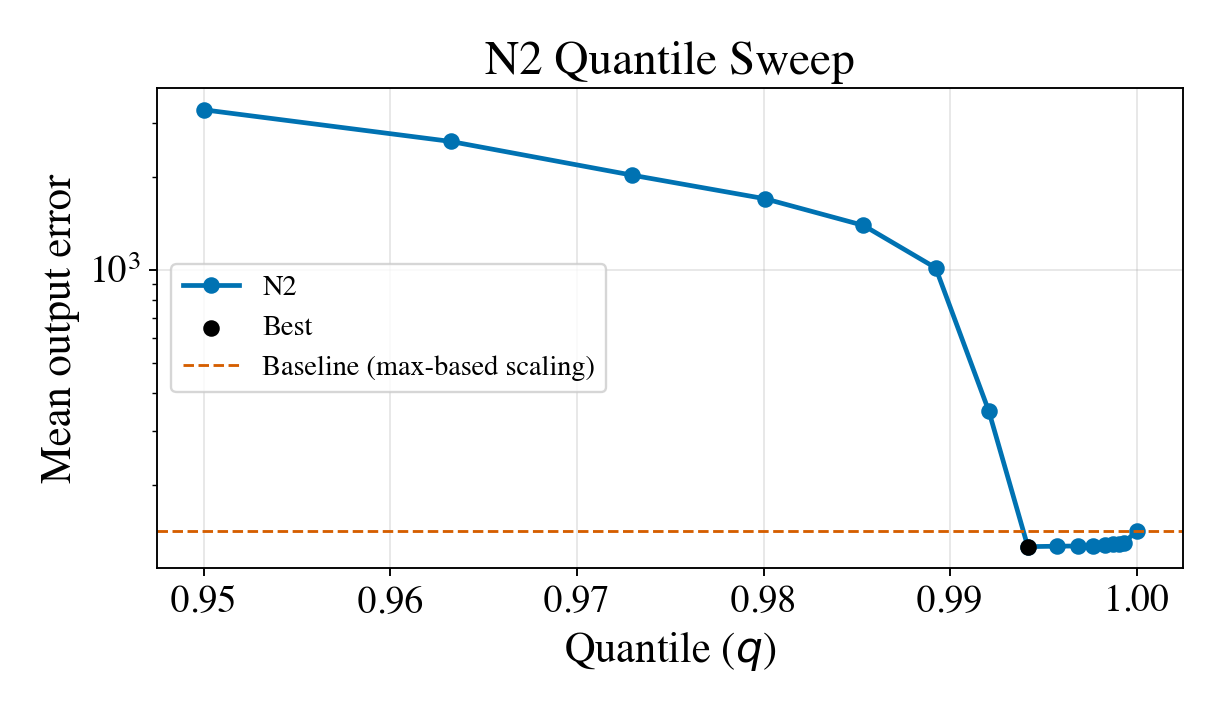}
    \caption{N2 quantile sweep over selected-layer mean linear output error.}
    \label{fig:n2_quantile_sweep}
\end{figure}

\subsection{Quantile Sweep (N2)}

Figure~\ref{fig:n2_quantile_sweep} shows the N2 quantile sweep results.
The best quantile is approximately 0.994, which corresponds to a high but not maximum quantile.
The best N2 point achieves an 11.084\% improvement over the selected-layer baseline.
With smaller quantiles, the error increases sharply, indicating that the estimate becomes too small and under-migrates activation values.
And as the quantile approaches 1, the error also increases but more gradually, indicating that the estimate becomes too large due to outliers and over-migrates activation values.
This experiment indicates that using a quantile-based scaling with \(q<1\) can reduce sensitivity to rare activation outliers and improve quantization error compared to max-based scaling.

\begin{figure}[t]
    \centering
    \includegraphics[width=\columnwidth]{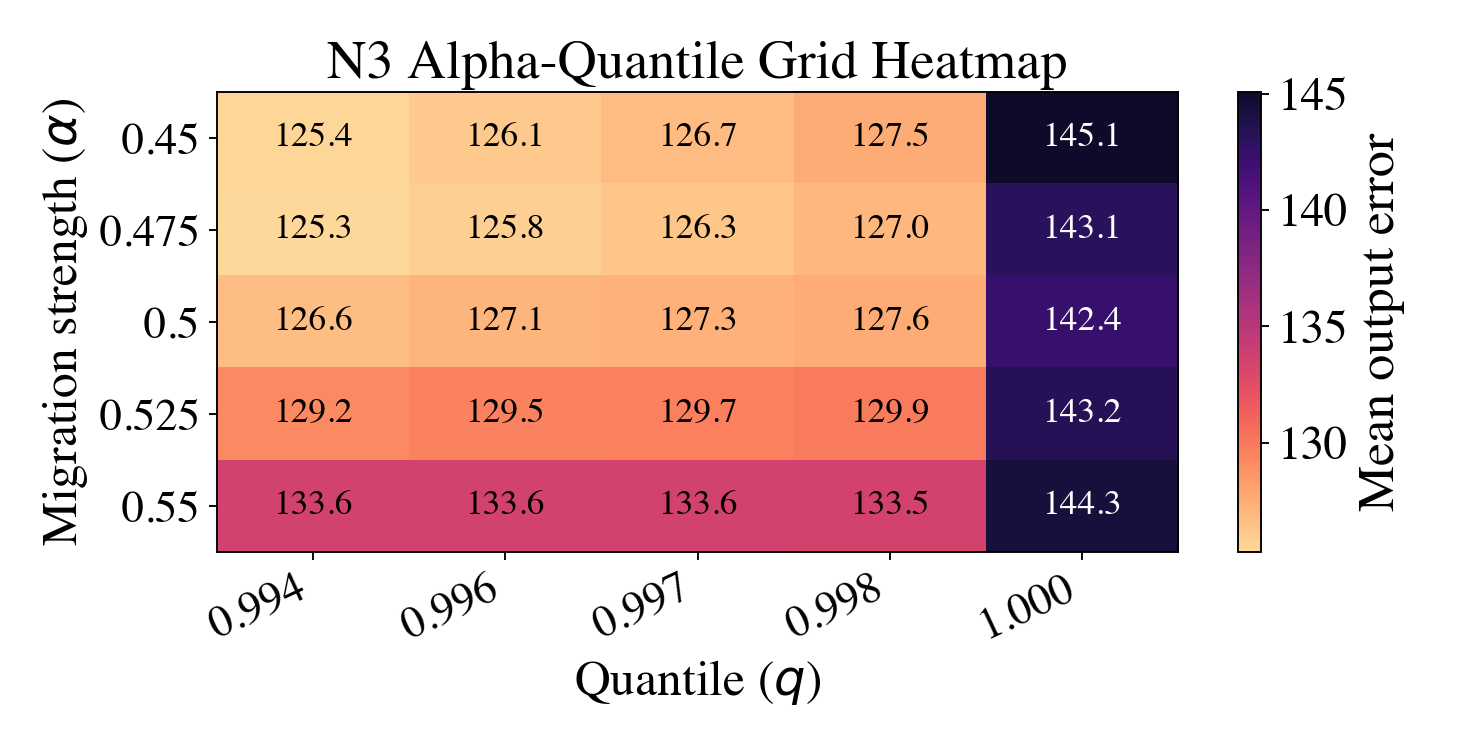}
    \caption{N3 alpha-quantile grid heatmap of selected-layer mean linear output error.}
    \label{fig:n3_alpha_quantile_heatmap}
\end{figure}

\subsection{Joint Alpha-Quantile Grid Search (N3)}

We visualize the N3 alpha-quantile grid search results as a heatmap in Figure~\ref{fig:n3_alpha_quantile_heatmap}.
The best point is at \((\alpha,q)=(0.475,0.994)\), which is close to the best points in N1 and N2, indicating that the optimal alpha and quantile values are somewhat independent.
The best N3 point achieves a 12.032\% improvement over the selected-layer baseline, which is a modest gain over the best N2 point, indicating that jointly tuning alpha and quantile can yield some additional improvement but the majority of the gain comes from the quantile-based scaling.

\begin{figure}[t]
    \centering
    \includegraphics[width=\columnwidth]{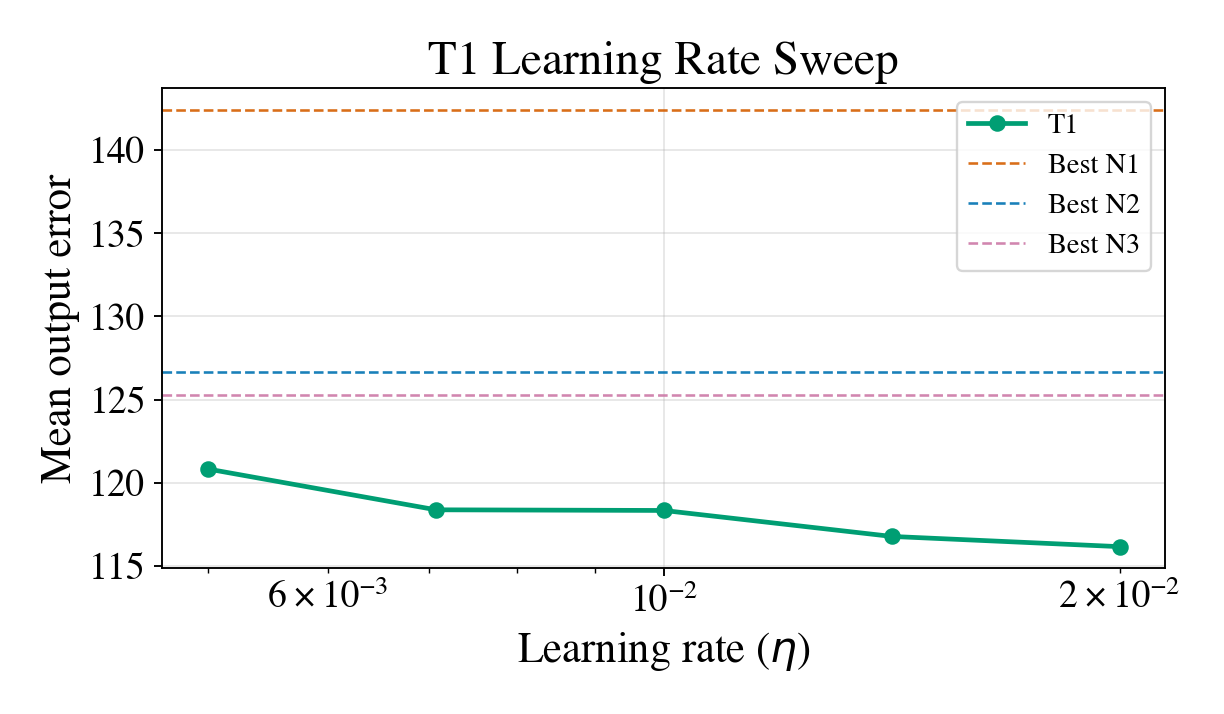}
    \caption{T1 learning-rate sweep compared to N-family best outcomes.}
    \label{fig:t1_lr_sweep_with_n_refs}
\end{figure}

\subsection{Training from Rotation-Only Initialization (T1)}

Figure~\ref{fig:t1_lr_sweep_with_n_refs} shows the T1 learning-rate sweep results compared to the best outcomes from the N families.
We find that as the learning rate increases, the error decreases. The best learning rate is 0.02, which achieves an 18.447\% improvement over the selected-layer baseline and a 7.165\% improvement over the best N3 point.
This experiment indicates that training the scaling factors can further reduce quantization error.

\begin{figure}[t]
    \centering
    \includegraphics[width=\columnwidth]{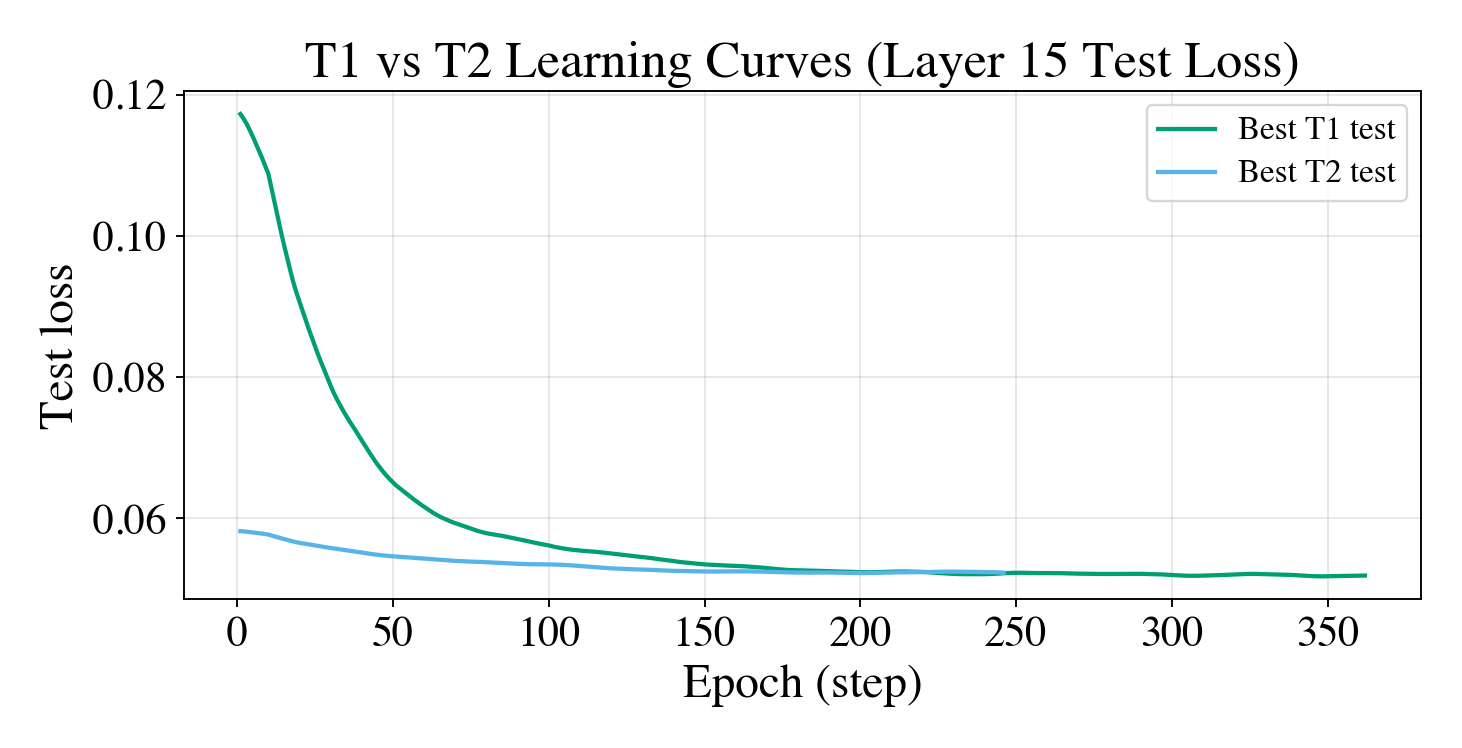}
    \caption{Comparison of T1 and T2 training curves on decoder-block 15 down-projection test loss (smoothened).}
    \label{fig:t2_layer15_test_loss_t1_vs_t2}
\end{figure}

\begin{figure}[t]
    \centering
    \includegraphics[width=\columnwidth]{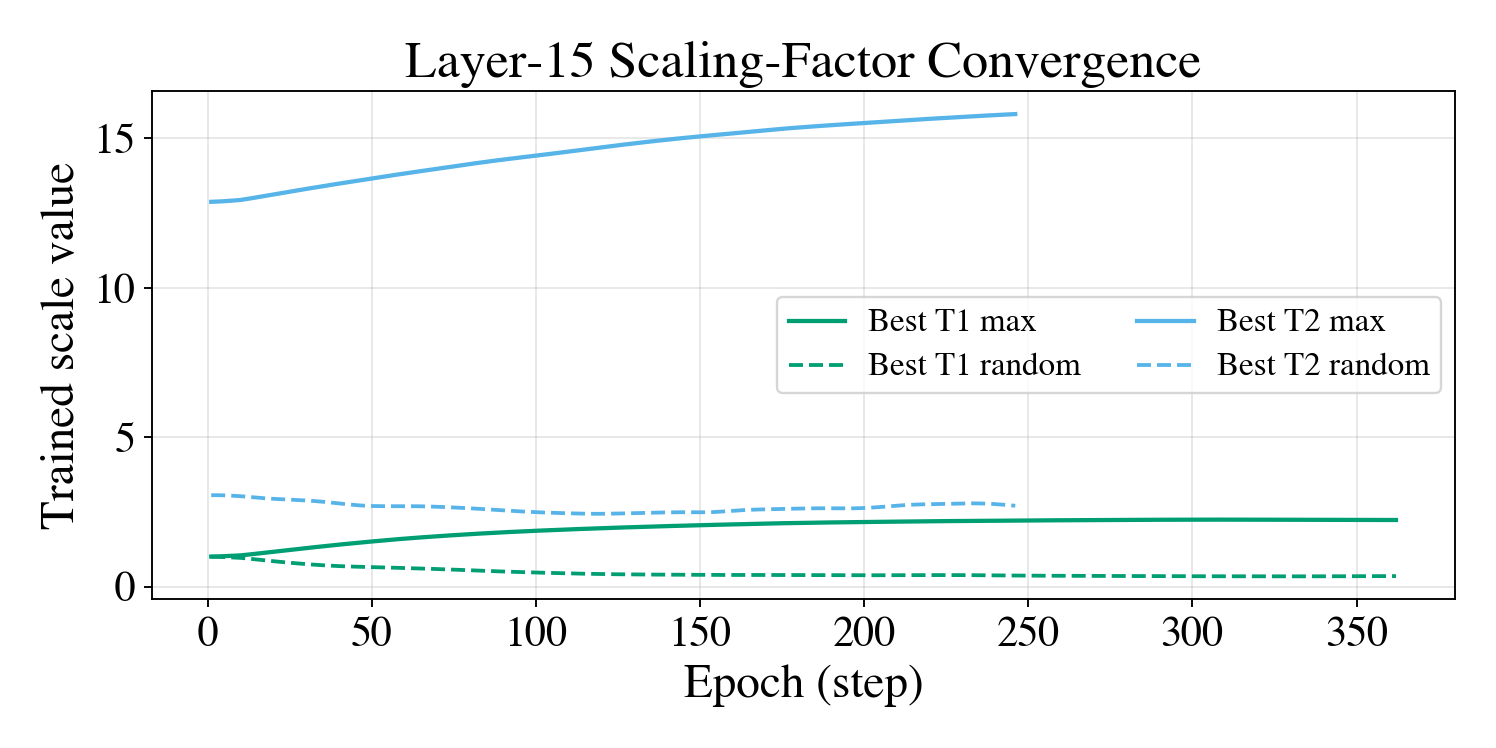}
    \caption{Layer-15 scaling-factor convergence under different initial values, contrasting T1 and T2 optimization behavior. Solid lines show scaling factors for the channels with the highest max absolute value, dashed lines correspond to a randomly selected channel.}
    \label{fig:layer15_scale_convergence_rerun}
\end{figure}

\subsection{Training from N3 Warmstart Policies (T2)}

The results of T1 prove that training can further reduce error, and the results of N3 provide good warmstart policies for T2.
The best T2 point is at \((\alpha,q,\eta)=(0.475,0.997,0.02)\), which achieves an 18.335\% improvement over the selected-layer baseline, which fails to improve over the best T1 point, indicating that the N3-informed warmstart does not provide additional benefit over rotation-only initialization when training is used.
Figure~\ref{fig:t2_layer15_test_loss_t1_vs_t2} compares the training curves of T1 and T2 on layer 15, which is the most critical layer. We see that both runs are stable and converge to similar final test loss values, with T2 starting from a significantly better initialization but T1 catching up after around 200 steps.
This suggests that the optimization landscape is relatively smooth and allows both runs to find good solutions, and that the N3-informed warmstart mainly provides a better starting point but does not lead to a better final solution compared to rotation-only initialization when training is used.
To understand the optimization behavior better, we also visualize the convergence of the scaling factor for layer 15 in T1 and T2 in Figure~\ref{fig:layer15_scale_convergence_rerun}.
We see that the two runs converge to completely different final scaling values, with T1 converging to much smaller values than T2. However, both runs share a common pattern of increasing the scaling factor for the channel with the highest max absolute value, while decreasing the scaling factor for the other, random channel.
In fact, we find a Pearson correlation of 0.98 between the scaling factors provided by T1 and T2 considering all channels, which indicates that both runs learn a very similar relative scaling pattern across channels, but with different overall magnitudes.
This indicates that the optimization landscape has a consistent structure that guides both runs to learn similar relative scaling patterns, but the overall scale can vary significantly depending on the initialization and optimization path taken.

\begin{figure}[t]
    \centering
    \includegraphics[width=\columnwidth]{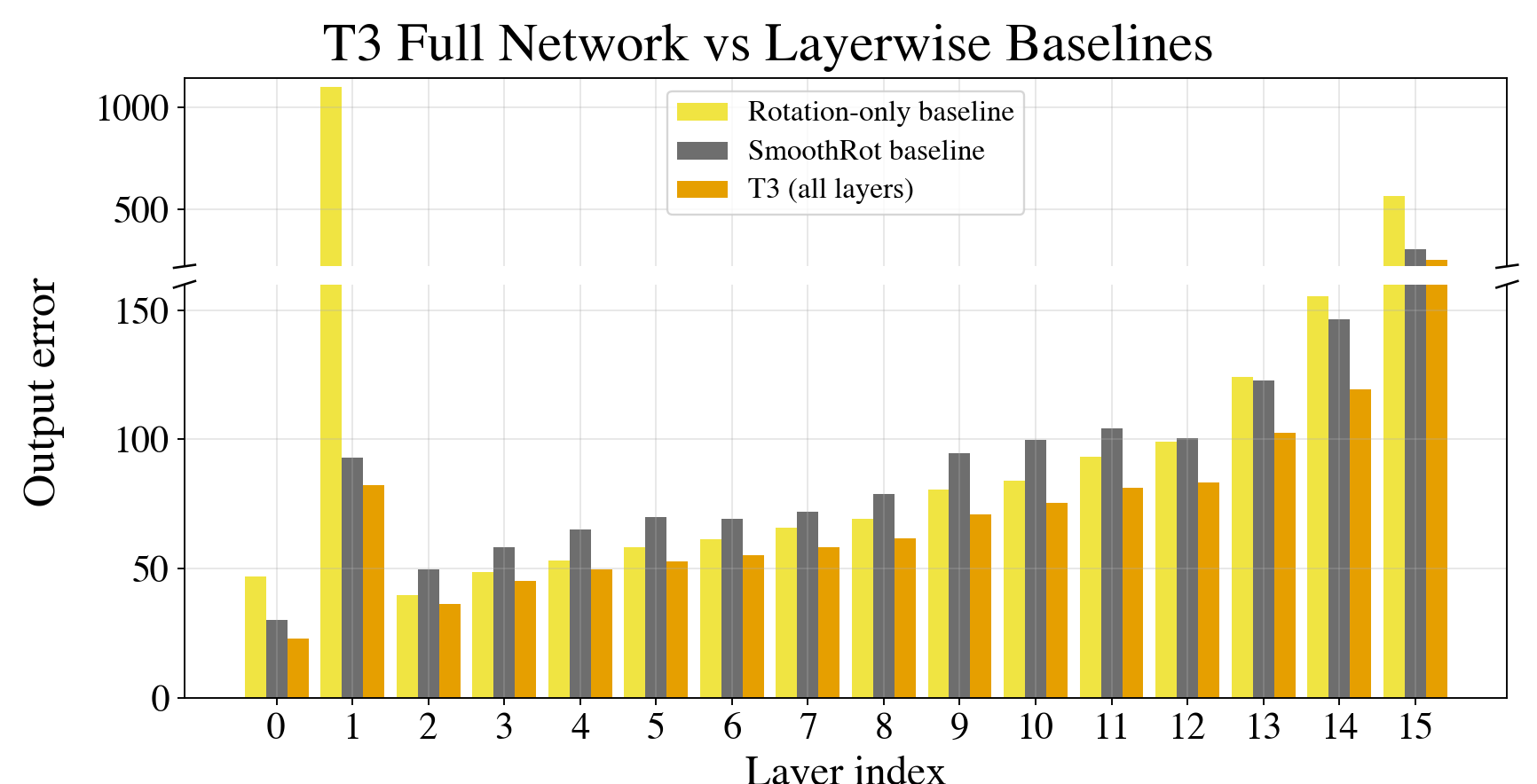}
    \caption{Layer-wise baselines versus T3 full-network replay comparison over all decoder-block down-projection layers.}
    \label{fig:layerwise_baseline_vs_t3_full_network}
\end{figure}

\subsection{Full-network replay (T3)}

T3 replays the best T2 policy on all decoder-block down-projection layers without changing \((\alpha,q,\eta,T,T_w)\). The resulting mean linear output error is 78.08 across 16 down-projection layers, compared with 97.51 for the SmoothRot baseline across the same set. This corresponds to an absolute gain of 19.43 and a relative gain of 19.93\%.
Role-conditioned analysis (using the selected-layer role mapping) shows gains in both groups: critical layers improve from 191.73 to 157.88 (17.66\%), and normal layers improve from 68.47 to 53.96 (21.12\%).
Figure~\ref{fig:layerwise_baseline_vs_t3_full_network} compares the layer-wise baselines with the T3 full-network replay results. We see that all down-projection layers improve over both baselines, with the most critical layers showing the largest absolute improvements, while the normal layers show more modest absolute improvements but larger relative improvements.
It is also interesting to note that while the original SmoothRot baseline achieves significantly better error on critical layers (0,1,13,14,15) compared to the rotation-only baseline, it falls behind on normal layers (2-12).
In contrast, our currently proposed method achieves consistent improvements across all down-projection layers, which suggests that the learned scaling factors can adapt to the specific characteristics of each layer and provide more balanced improvements compared to previous heuristics.

\section{Discussion and Limitations}

\label{sec:discussion}

The results support three technical points.
First, quantile policy matters more than alpha-only adjustment in the no-training regime: N2 and N3 provide clear gains while N1 does not.
Second, constrained training on policy-informed candidates improves the selected-layer objective substantially over no-training baselines.
Third, replaying the best selected-layer policy on all down-projection layers yields the strongest overall full-layer gain, indicating that selected-layer optimization can transfer when policy consistency is enforced.

These outcomes also clarify the mechanism behind prior failures: max-based calibration can overestimate activation scale in the presence of rare spikes, which induces over-migration and harms balance between activation and weight quantization.
Replacing max with a high quantile mitigates this effect while preserving sensitivity to dominant channels, and training then refines the relative channel pattern further.

The study has important limitations.
First, experiments are restricted to a single model (LLaMA-3.2-1B), and one dataset split (Wikitext-2);
broader validation on larger models, downstream tasks, and alternative calibration data is still needed.
Second, optimization is performed on linear output error rather than direct perplexity/task loss, so alignment with end-task quality is inferred but not explicitly optimized.
Finally, training-time overhead and deployment trade-offs were not the primary target of this work and should be analyzed in future system-level evaluations.

\section{Conclusion}

This paper examined a specific but consequential limitation of SmoothRot-style PTQ: degradation caused by over-migration when scale calibration relies on max-dominated activation statistics.
We introduced a quantile-robust scaling policy and evaluated it in a policy-structured pipeline that separates no-training search from constrained scale learning.
In the no-training regime, quantile-based policies produced clear gains while alpha-only tuning did not.
With training, learned scales yielded the strongest selected-layer improvements.
Replaying the best policy on all layers improved the mean error from 97.51 to 78.08 (19.93\%), showing that selected-layer findings can transfer when policy consistency is maintained.
Overall, the results indicate that robust migration control plus lightweight scale optimization is an effective and practical direction for improving low-bit activation quantization in LLMs.

\section*{Acknowledgment}
The authors would like to thank the Doctoral School of Applied Informatics and Applied Mathematics of Obuda University for their valuable support.

\printbibliography

\end{document}